\documentclass[sigconf,nonacm]{acmart}

\AtBeginDocument{%
  \providecommand\BibTeX{{%
    \normalfont B\kern-0.5em{\scshape i\kern-0.25em b}\kern-0.8em\TeX}}}

\renewcommand\footnotetextcopyrightpermission[1]{} 
\usepackage{times}
\usepackage{epsfig}
\usepackage{graphicx}
\usepackage{amsmath}
\usepackage{amssymb}
\usepackage{color,xcolor}
\usepackage{multirow}
\usepackage{setspace}
\usepackage{mathrsfs}
\usepackage{subcaption}
\usepackage{amsmath} 

\newcommand{\etal}{\textit{et al.}}
\usepackage{fancyhdr}
\usepackage[margin=0pt,font=small,labelfont=bf,labelsep=endash,tableposition=top]{caption}
\begin{document}

\title{DiffCalib: Reformulating Monocular Camera Calibration as Diffusion-Based Dense Incident Map Generation}


\author{Xiankang He}
\authornote{Both authors contributed equally to this research.}
\affiliation{%
  \institution{Zhejiang University of Technology}
  \city{Hangzhou}
  \country{China}}
\email{201806022509@zjut.edu.cn}

\author{Guangkai Xu}
\authornotemark[1]
\orcid{0000-0001-8783-8313}
\affiliation{%
  \institution{Zhejiang University}
  \city{Hangzhou}
  \country{China}}
\email{guangkai.xu@gmail.com}

\author{Bo Zhang}
\affiliation{%
\institution{Zhejiang University}
\city{Hangzhou}
\country{China}}
\email{zhangboknight@gmail.com}

\author{Hao Chen}
\affiliation{%
\institution{Zhejiang University}
\city{Hangzhou}
\country{China}}
\email{stanzju@gmail.com}

\author{Ying Cui}
\affiliation{%
\institution{Zhejiang University of Technology}
\city{Hangzhou}
\country{China}}
\email{cuiying@zjut.edu.cn}

\author{Dongyan Guo}
\authornote{Dongyan Guo is the corresponding author.}
\affiliation{%
\institution{Zhejiang University of Technology}
\city{Hangzhou}
\country{China}}
\email{guodongyan@zjut.edu.cn}


\begin{abstract}
Monocular camera calibration is a 
key precondition
for numerous 3D vision applications. Despite considerable advancements, existing methods often hinge on specific assumptions and struggle to generalize across varied real-world scenarios, and 
the performance is limited by
insufficient training data. 
Recently, diffusion models trained on expansive datasets have been 
confirmed
to maintain the capability to generate diverse, high-quality images. This success suggests a strong potential of the models to effectively understand varied visual information.
In this work, we leverage the comprehensive visual knowledge embedded in pre-trained diffusion models to 
enable more robust and accurate monocular camera intrinsic estimation.
Specifically, we reformulate the problem of estimating the four degrees of freedom (4-DoF) of camera intrinsic parameters as a dense incident map generation task. The map details the angle of incidence for each pixel in the RGB image, and 
its format
aligns well with the paradigm of diffusion models. 
The camera intrinsic then can be derived from the incident map with a simple non-learning RANSAC algorithm during inference.
Moreover, to further enhance the performance, we jointly estimate a depth map to provide extra geometric information for the incident map estimation.  
Extensive experiments on multiple testing datasets demonstrates
that our model achieves state-of-the-art performance, 
gaining
up to a 40\% reduction in prediction errors. 
Besides, the experiments also show that the precise camera intrinsic and depth maps estimated by our pipeline can greatly benefit practical applications such as 3D reconstruction from a single in-the-wild image.

\end{abstract}

\ccsdesc[500]{Computing methodologies}
\ccsdesc[500]{Computing methodologies~Camera calibration}

\keywords{Monocular Camera Calibration; Depth Estimation; Generative Model}
\begin{teaserfigure}
  \includegraphics[width=\textwidth]{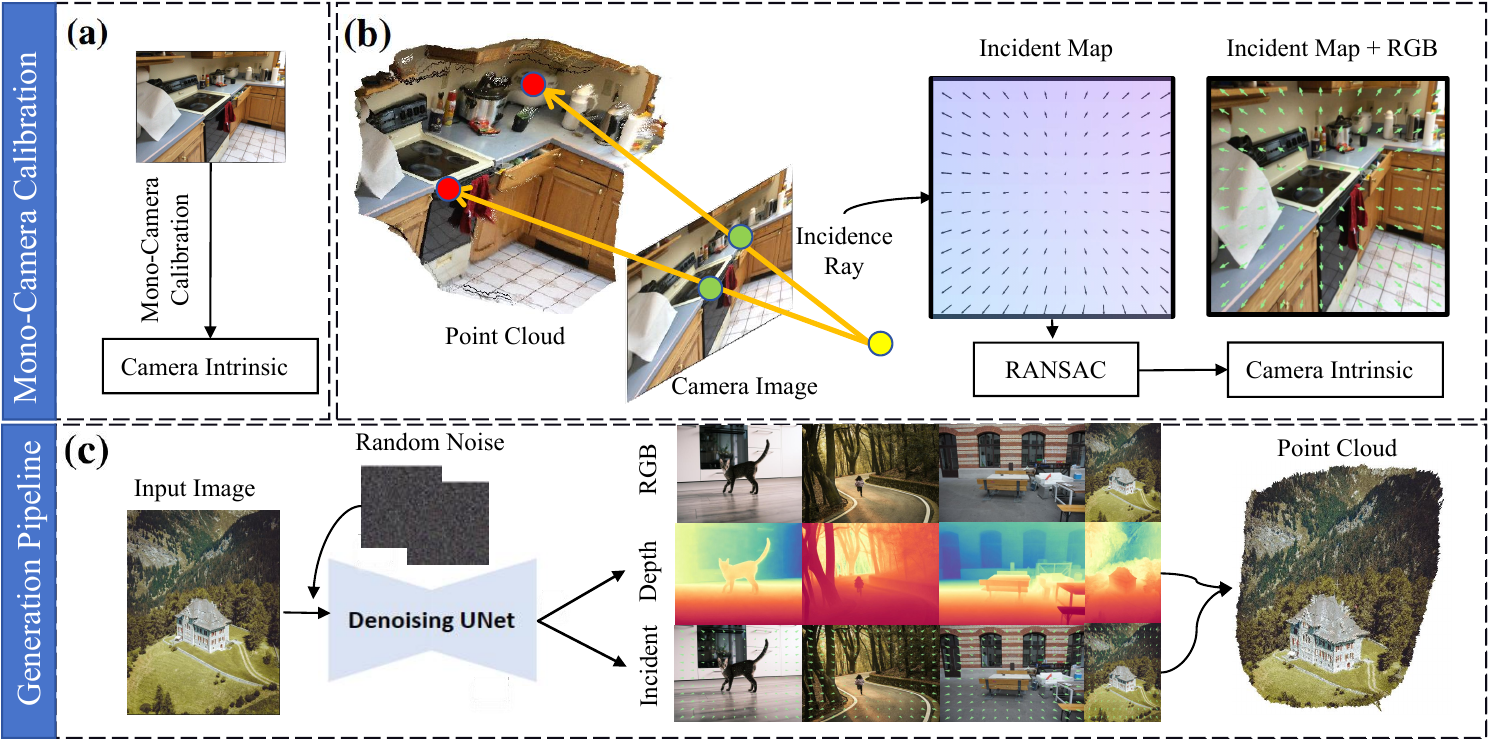}
  \caption{We reformulate monocular camera calibration as a diffusion-based incident map generation task. (a) Our pipeline enables robust camera intrinsic estimation from a single image. (b) The `incidence map' represents incident rays pointing from the camera image pixels to the 3D point cloud. The camera intrinsic can be derived from the incident map with the RANSAC algorithm. (c) Our pipeline leverages the paradigm of latent diffusion models, takes the RGB image and random Gaussian noise as input, and generates the incidence map and depth map together. Subsequently, in-the-wild 3D reconstruction is enabled with the predicted depth and intrinsic.}
  \label{fig:gencamera}
\end{teaserfigure}


\maketitle


\section{Introduction}

Monocular camera calibration \cite{Wildcamere, jin2022perspective, hold2018perceptual, lee2020neural} aims to estimate the intrinsic properties of a camera from a single image, which is important for many downstream tasks of 3D scene reconstruction and understanding. 
While the robustness and accuracy of monocular camera calibration are receiving increasing attention, existing approaches either rely on geometric principles like Manhattan world assumption ~\cite{coughlan1999manhattan}, or are based on specific objects like checkerboards or human faces~\cite{hu2023camera}, and therefore can hardly generalize to diverse real scenarios. 


To alleviate the reliance mentioned above, Zhu \etal~\cite{Wildcamere} proposed to estimate camera intrinsic without geometric or object assumption. 
They introduce an innovative concept termed as incident field, which delineates the direction between the 3D point cloud in the world location and the camera's optical center. After training, the camera intrinsic can be recovered robustly with a simple RANSAC ~\cite{fischler1981random} algorithm. Compared with previous methods, the incident field reformulates the camera intrinsic as a pixel-wise map and achieves promising generalization. However, just like the ill-pose property of monocular depth estimation, the monocular camera calibration struggles to estimate results accurately with confidence from a single image, 
mainly
due to the limited training data.

Recently, a series of advanced approaches~\cite{ke2023repurposing,xu2024diffusion,fu2024geowizard} for monocular depth estimation have emerged, which leverage the robust knowledge priors embedded within Stable Diffusion models. Through strategic fine-tuning protocols, these methods have demonstrated exceptional capabilities in achieving commendable zero-shot generalization. 


In this paper, motivated by the successful visual knowledge transfer from image generation to depth estimation, we propose to solve the monocular camera calibration problem by reformulating it as an incident map generation task. The incident map details the angle of incidence for each pixel in the RGB image and its format aligns well with the paradigm of diffusion models. Thus, it can be learned by properly fine-tuning and enforcing the Stable Diffusion models.
With such design, we can effectively address the community's focus on the robustness and accuracy by leveraging the advantages of diffusion models. 
Specifically, we freeze the VAE encoder and decoder of Stable Diffusion, and fine-tune the U-Net to learn the noise added to the incident map. It can bring two main advantages: 1) We prove that the rich visual information of Stable Diffusion models can benefit not only perception tasks but also the camera characteristics estimation. 2) By regarding the estimation process as a probabilistic one, the confidence of the predicted incident map will be increased by de-noising from different noise maps and ensembling the corresponding results. 

Besides, the connection between camera intrinsic parameters and depth maps in single-image 3D reconstruction, as analyzed by Yin \etal~\cite{yin2020learning}, highlights the inherent relationship between these two elements. To further enhance the performance, we incorporate depth information into our pipeline. The network is enforced to concurrently estimate both the incident map and the depth map of an input RGB image, which is proved to improve results in both modalities according to our experiments. 

Using the estimated incident map, the camera's intrinsic parameters can be accurately derived through the non-learning-based RANSAC~\cite{fischler1981random} algorithm. Our extensive quantitative and qualitative experiments demonstrate that our method outperforms recent approaches and achieves state-of-the-art results. Furthermore, our comprehensive ablation studies 
verify
the effectiveness of each component. Additionally, by leveraging the estimated depth map, we can project the 2D image into 3D space, facilitating 3D reconstruction from a single in-the-wild image.

Overall, our contributions can be summarized as follows.
\begin{itemize}
    \item To the best of our knowledge, this is the first work to leverage the visual knowledge priors of diffusion models to reformulate the camera intrinsic estimation as the task of generating a dense incident map. This approach significantly improves the robustness and accuracy of the estimates.
    \item We introduce a method to jointly estimate the incident map and the depth map, leveraging their intrinsic relationships to enhance the performance of both.
    \item Utilizing the predicted depth map and camera intrinsics derived from the incident map, our approach can benefit downstream applications such as 3D reconstruction from a single image, even in challenging in-the-wild scenes.
\end{itemize}

\begin{figure*}
	\centering 
	\includegraphics[width=1\textwidth]{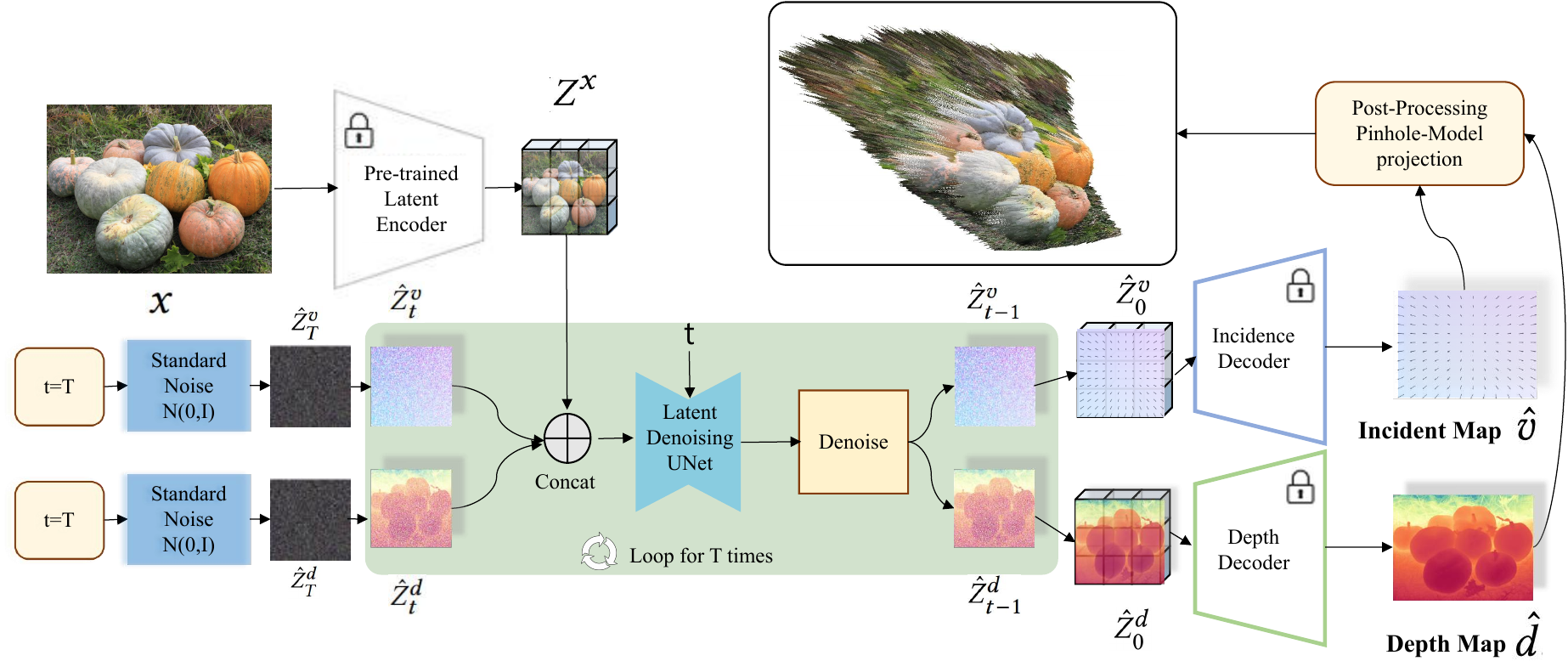}
	\caption{Overview of the generation pipeline. Given an image $x$,  we generate the incident map $\hat{v}$ and depth map $\hat{d}$ using the denoising U-Net from two randomly sampled Gaussian noises $\hat{Z}_T^v$ and $\hat{Z}_T^d$. The generated $\hat{v}$ and $\hat{d}$ are projected into 3D space to recover the 3D scene shape. It is worth mentioning that the denoising process in the green part loops for $T$ times. Please see more details in \S ~\ref{section:C}.}
	\label{fig-framwork}
\end{figure*}

\section{related work}
\subsection{Monocular Camera Calibration}
Monocular camera calibration is a crucial process in computer vision, with a significant emphasis on geometric considerations and object properties. Within the realm of research, scholars employ various methodologies to tackle this challenge. 
Early methods
are rooted in geometric principles, such as the Manhattan world assumption~\cite{coughlan1999manhattan}, which leverages specific geometric regularities within scenes to infer internal camera parameters. 
Later, researchers try to
utilize distinct objects or patterns, such as chessboard~\cite{zhang2000flexible} or line segments~\cite{von2008lsd, akinlar2011edlines}, for calibration purposes. 
The emergence of these methodologies has enriched the field of camera calibration, offering a diverse range of approaches for estimating intrinsic parameters. 
However, the dependent upon specific assumptions make these methods lead to poor generalization ability across varied real-word scenarios.
Recently, researchers have explored the use of real-world objects, including faces~\cite{hu2023camera} or other real-world items~\cite{grabner2019gp2c, chen2019bpnp, sturm2005multi}, for camera calibration, thus expanding the scope and versatility of calibration techniques.
However, they still face the problem of specific objects requirement like checkerboards or human faces, which limit their generalization to diverse in-the-wild scenarios.
In contrast, our method reformulate the calibration problem as an image generation task. By leveraging the advantages of diffusion model, our method can effectively predict the four degrees of freedom (4-DoF) of camera intrinsic parameters with only one single undistorted image, which discards conventional assumptions and specific objects requirements.

\subsection{Learnable Monocular 3D Priors}

The most well-known 3D priors in computer vision are monocular depth~\cite{yin2020learning, ranftl2021vision, Ranftl2022, yin2023metric3d, xu2022towards, xu2024diffusion} and surface normal~\cite{bae2021estimating, xu2024diffusion}. These priors are fundamental for the transition from 2D to 3D tasks, as they provide essential information about the depth and orientation of surfaces in a scene. Recently, Jin \textit{et. al.} ~\cite{jin2023perspective} introduce the perspective field for single image camera calibration, which offers per-pixel information about the camera view. It is parameterized by an up vector and a latitude value, enabling a representation that capturing local perspective properties in an image. This perspective field proves to be effective for single image camera calibration tasks.
In a parallel development, Zhu \textit{et. al.} ~\cite{Wildcamere} introduce the incidence field that defined as the incidence rays between 3D points and 2D pixels to estimate the camera intrinsic via exploiting the monocular 3D prior. 
A Transformer-based fully-connected CRFs neural network~\cite{yuan2022newcrfs} is utilized for incidence field estimation.
Compared with the perspective field approach ~\cite{jin2023perspective} that requires panorama images as training data, this incidence field only need undistorted images, which demonstrates a valuable solution to in-the-wild monocular camera calibration.    
However, its performance is limited by insufficient training data for the neural network.
To solve the problem, our approach further formulate the incidence field estimation as an incidence map generation task, thus leverage 
the powerful knowledge priors embedded within the generative Stable Diffusion models to achieve more generalized and robust in-the-wild single image camera calibration. 

\subsection{Diffusion Models}

The Diffusion Denoising Probability Model (DDPM), commonly referred
as the Diffusion Model~\cite{ho2020denoising_ddpm}, presents a novel generative paradigm distinct from Generative Adversarial Networks (GANs)~\cite{goodfellow2014generative}. Renowned for its exceptional generation quality and controllable synthesis, DDPM is gaining popularity across diverse domains. 
The core of the model 
involves training a denoising encoder to execute the reverse process of Markov diffusion~\cite{song2020denoising_ddim}. 
DDPM demonstrates commendable performance in both unconditional and conditional generation tasks~\cite{dhariwal2021diffusion, chen2022generalist}, as well as image-to-image translation~\cite{saharia2021image}.
Recent advancements have extended the utility of the Diffusion Model to generate high-resolution synthetic images~\cite{rombach2022high} with the more powerful latent diffusion model. The model conducts diffusion in lower-resolution spaces and effectively reduce computational cost while maintaining high fidelity in synthetic image generation.
Our approach capitalizes on the robust priors embedded within the diffusion model to generate high-precision incident maps and depth maps, which enables more accurate intrinsic parameter estimation and facilitates in-the-wild dense 3D reconstruction.

\subsection{3D Scene Reconstruction}
Conventional researches mainly focus on
recovering individual objects from single images~\cite{barron2020shape, wang2018pixel2mesh, wu2018learning}, such as cars, tables, and humans. However, these methods are tailored to specific object classes and with poor generalization performance, 
making them unsuitable for real-world scene reconstruction task.
Recently, Yin \textit{et. al.} ~\cite{Wei2021CVPR_leres} proposes a method named LeReS that enables the capability of recovering the 3D scene shape from a single image, and Xu \textit{et. al.} \cite{xu2023frozenrecon} optimizes 3D scene shape from an input video by supervising multi-view consistency. 
However, they either lack the robustness brought by strong pre-trained backbones or rely on the video input paradigm.
Our approach can predict precise depth maps and camera intrinsic parameters from a single image, which can benefit the downstream application such as achieving superior 3D shape reconstruction.

\section{Method}
The overall pipeline of DiffCalib is shown in Figure ~\ref{fig-framwork}. Our approach further formulates the
incidence field estimation as an incidence map generation task(\S \ref{section:C}), with rich visual knowledge of the pre-trained diffusion models leveraged to enhance the robustness of calibration. Additionally, the simultaneous estimation of depth maps and incident maps facilitates performance boosting and applications like 3D reconstruction from a single image (\S \ref{sec: 3d_recon}). To begin with, let's introduce the concepts of the incident map (\S \ref{section:A}), and reformulate it as diffusion-based generation (\S \ref{section:B}).

\subsection{Incident Map}
\label{section:A}
The incident map contains the collection of incident rays originating from points within the scene and passing through corresponding pixels on the camera's imaging plane.
It outlines the array of rays extending from each pixel position on the imaging plane to the camera's focal point.

Take the pinhole camera model as an example. 
Mathematically, for any pixel $p$ of coordinate $(x, y)$ on the imaging plane, the incident map vector $v$, denoted as $V(p)$, can be expressed as follows.
\begin{equation}
v = V(p) = V(x, y) = [\frac{x - b_x}{f_x}, \frac{y - b_y}{f_y}, 1]^T \label{eq:vv}
\end{equation}
Where $b_x$ and $b_y$ denote the optical center location along the x-axis and y-axis, and
$f_x$ and $f_y$ represent the focal length along the x-axis and y-axis, respectively. 


The incident map emerges as a crucial 3D prior due to its ability to convey essential information regarding the camera's viewing perspective. 
In contrast to directly estimating the 4-DoF camera intrinsic parameters, the dense incident map bears a closer resemblance to natural images and is invariant to image transformations such as cropping and resizing. This similarity enables it to potentially leverage the extensive knowledge priors of networks that have been pre-trained on a diverse array of real-world scenes.




\begin{figure}
	\centering 
	\includegraphics[width=1\columnwidth]{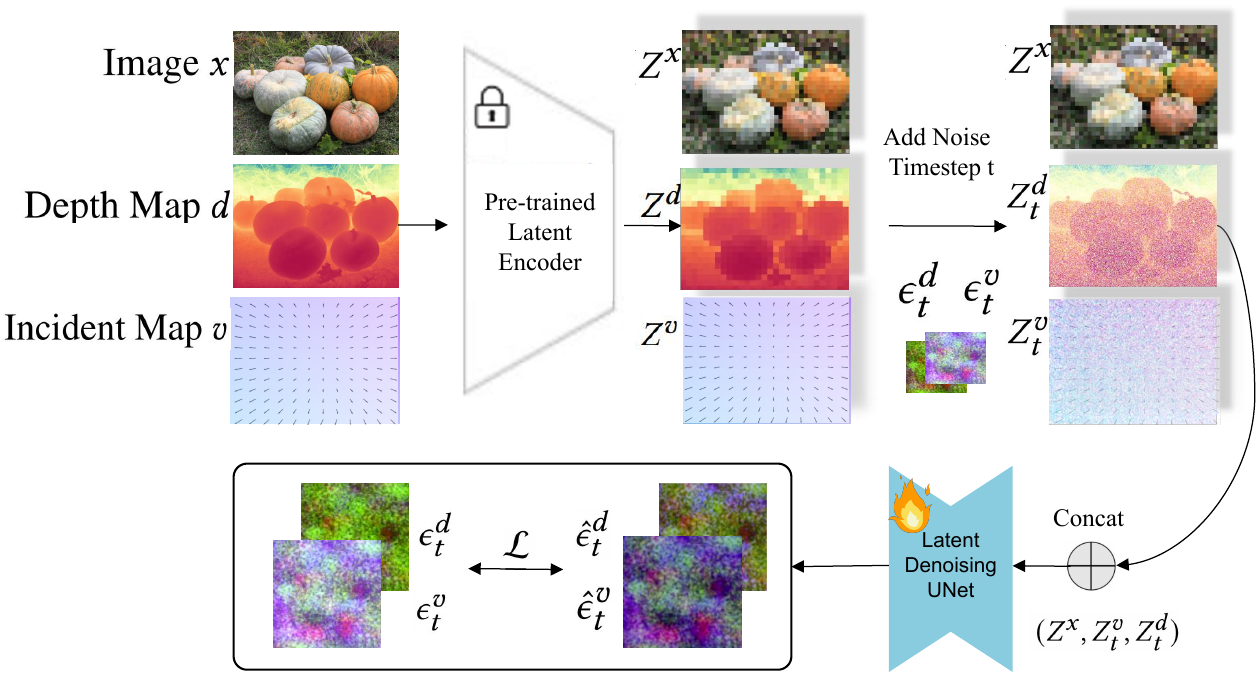}
	\caption{Overview of the training pipeline. We freeze the latent encoder and encode the input image $\mathbf{x}$, incident map $\mathbf{v}$, and depth map $\mathbf{d}$ into the latent space. Then, the U-Net is trained to predict the noise added to the depth and incident map latent codes, denoted as $\hat{\epsilon}_t^d$ and $\hat{\epsilon}_t^v$, respectively. Loss function is computed between the estimated noise and the added ground-truth noise.
 }
	\label{fig-train}
\end{figure}
\subsection{Diffusion-Based Generation}
\label{section:B}
With the dense representation of the incident map, we reformulate the monocular camera calibration task as an incident map generation process by leveraging the rich knowledge priors of the Stable Diffusion v2.1 model. The Stable Diffusion is composed of a VAE autoencoder that transfers images to the latent space, and a U-Net that estimates the noise added to the image.

During training, 
the incident map is firstly encoded to the latent space as $Z^v$ with the pre-trained VAE encoder. Then, we gradually add random sample Gaussian noise to get the noising incident latent codes
$Z_t^v$:
\begin{align}
Z_t^v &= \sqrt{\bar{\alpha}_t} Z_0^v + \sqrt{1 - \bar{\alpha}_t} \epsilon \label{eq:vt_add}
\end{align}
where $Z_0^v = Z^v$ is the initial incident map and $\bar{\alpha}_t$ is noise scheduler that controls sample quality. The timestep $t \sim (1, T)$ and $\epsilon \sim \mathbf{N}(0, I)$.
Then, the noisy encoded incident map passes through the U-Net $\epsilon_\theta$ to predict the estimated noise $\hat{\epsilon}$.
The model is trained by minimizing the $L_2$ loss of  the estimated noise $\hat{\epsilon}$ and the ground-truth added noise $\epsilon$

During inference, we denoise the input $Z_t^v$ to $Z_{t-1}^v$ step by step from $Z_T^v$ to $Z_0^v$ with the trained denoiser U-Net. Finally, the estimated incident map is recovered from $Z_0^v$ with the pre-trained VAE decoder.

\subsection{DiffCalib}
\label{section:C}

Our approach is based on the pre-training of 2D latent stable diffusion, also known as Stable Diffusion v2.1~\cite{rombach2022high}. Our objective is to further formulate the incidence field estimation as an incidence map generation task, thus leveraging the powerful knowledge priors embedded within the generative Stable Diffusion models to achieve more generalized and
robust in-the-wild single-image camera calibration. However, we encounter a challenge in this transition process. The structural distribution of the incident map is typically uniform, comprising only four parameters that are distinct from the image training during pre-training. In fact, there exists a similarity between pixels and their values, posing difficulty for the diffusion model to effectively utilize the priors of image generation. Bridging the gap between the diffusion model generation paradigm and our target image incident map is therefore crucial.

\paragraph{Enhance Incident Map Generation by Jointly Learning with Depth Map.}

Spired by Yin \etal \cite{yin2020learning} which proposes a method that can estimate the shift and focal length from the 3D point cloud which are projected from the depth map and a random initial focal length. 
This indicates the association between depth information and camera intrinsic parameters, highlighting the correlation between the incident map and depth map. With the assistance of the depth map, we can map the rays of incidence to the latent 3D space. Similarly, the incident map can also project the depth map from its 1D values into 3D space. Therefore, we jointly incorporate the depth and incident map into our generation model to improve the incident map generation.

Next, we will introduce our training pipeline in Figure~\ref{fig-train}.
When we input the image $x$ and incident map $v$, the paired depth map $d$ is also included. The depth map will be replicated into three channels, resembling RGB, to emulate the image input during the pre-training phase.

Then, the frozen VAE Encoder $E(\cdot)$ will encode the input x, d and v as $Z^x = E(x)$, $Z^d = E(d)$, $Z^v = E(v)$. And then $Z^d$, $Z^v$ will be added multi-resolution noises~\cite{whitaker2023multi-resolution} $\epsilon_t^d$ and $\epsilon_t^v$ to $Z_t^d$ and $Z_t^v$ with the timestep $t$, respectively. Along with the latent representation of the image $Z^x$, we concatenate these representations along the feature dimension $(Z^x, Z_t^v, Z_t^d)$.

Due to the modification, the input feature dimension will have 12 channels, whereas the U-Net input channel of Stable Diffusion v2.1 is originally 4. To accommodate this change, we triple the input channel and adjust the weight tensor of the input layer accordingly. Specifically, we duplicate the weight tensor of the input layer three times and divide its values by three to prevent inflation of activation magnitudes in the first layer.

Rethinking our generation paradigm, especially with the association between depth information and incident maps which can map the rays of incidence to the latent 3D space,
there we consider the the joint of incident map latent codes $Z_t^v$ and depth latent codes $Z_t^d$ can be seen as a 3D representation $Z_t^P$, allows the denoiser to effectively utilize the joint latent variable representation for denoising each modality. Thus,  we consider the denoise function as:
\begin{equation}
\hat{\epsilon}_t^P = \epsilon_\theta(Z^x, Z_t^P, t)
\end{equation}
where $\hat{\epsilon}_t^P$ can be split to $\hat{\epsilon}_t^v$ and $\hat{\epsilon}_t^d$ as the predicted noise of incident map and depth, separately. So the loss can be summarized as: 
\begin{align}
\begin{aligned}
\mathcal L = \mathbb{E}_{x,v,d,\epsilon_t^v \sim \mathcal{N}(0,I), t \sim U(T)} \| \epsilon_t^v - \hat{\epsilon}_t^v \|_2^2  + \\
\mathbb{E}_{x,v,d,\epsilon_t^d \sim \mathcal{N}(0,I), t \sim U(T)} \| \epsilon_t^d - \hat{\epsilon}_t^d \|_2^2
\end{aligned}
\end{align}



\paragraph{Inference Performance Improvement with the Ensemble Process.}

During the inference phase, with frozen VAE encoder to encode image $x$ to the latent code $Z^x$. 
we only input the image $x$, 
At this time, we initialize the ensemble-size noises $\hat{Z}_T^v$ and $\hat{Z}_T^d$ from the noise generator for the incident map and depth, separately. Each noise in the ensemble is paired with the aforementioned image, ensuring diversity across the noise samples.

By averaging the generated incident noise, where less accurate pixels undergo multiple generations to improve credibility, while accurate pixels exhibit greater stability, we have utilized this method to enhance the accuracy of the results and achieve greater stability in generating more reliable incident maps.
Then, with the input of $(Z^x, \hat{Z}_T^v, \hat{Z}_T^d)$, we put it into U-Net for denoising with multistep. Finally, we can obtain $\hat{Z}_0^v$ and $\hat{Z}_0^d$ which will be decoded by frozen VAE Decoder and we can get the ensemble incident maps $\hat{v}$ and depth maps $\hat{d}$.
Notely, the ensemble $\hat{v}$ and $\hat{d}$ will still be averaged by ensemble-size, and get the final incident map and depth map.
It's worth noting that while the incident map is reconstructed as is, the depth map transforms. Initially reconstructed into three channels, an averaging process is subsequently applied to condense it into a single channel.


\paragraph{Monocular Intrinsic Calibration from Incident Map.}
With the reconstructed incident map $\hat{v}$, we use a RANSAC method without assumptions to recover the camera's intrinsic matrix $K$. from the incident map. By leveraging the relationship between the incidence vector $v$ and the camera intrinsic parameters, the intrinsic matrix can be directly inferred.

With the 2D pixel location of the image $\mathbf{p} = [x, \ y, \ 1]^T$ and camera intrinsic $K$:
\begin{equation}
\mathbf{K} = \begin{bmatrix}
    f_x & 0 & b_x \\
    0 & f_y & b_y \\
    0 & 0 & 1
\end{bmatrix}
\end{equation}
, we can randomly sample two incidence vectors:
\begin{align}
    \mathbf{v}_1 = \mathbf{K}^{-1} \begin{pmatrix} x_1 \\ y_1 \\ 1 \end{pmatrix} = \begin{pmatrix} \frac{x_1 - b_x}{f_x} \\ \frac{y_1 - b_y}{f_y} \\ 1 \end{pmatrix} , 
    \mathbf{v}_2 = \mathbf{K}^{-1} \begin{pmatrix} x_2 \\ y_2 \\ 1 \end{pmatrix} = \begin{pmatrix} \frac{x_2 - b_x}{f_x} \\ \frac{y_2 - b_y}{f_y} \\ 1 \end{pmatrix} \label{eq:v1_2}
\end{align}
Then the camera's intrinsic matrix is estimated using the RANSAC algorithm and a minimal solver follows WildCamera ~\cite{Wildcamere}. Based on the relationship between the incidence vector and the camera's intrinsic parameters, we can directly derive the focal lengths as well as the pixel coordinates of the optical center:
\begin{align*}
\begin{aligned}
f_x &= \frac{x_1 - x_2}{v_{x}^1 - v_{x}^2} , & b_x &= \frac{1}{2}(x_1 - v_{x}^{1}f_x + x_2 - v_{x}^{2}f_x) \\
f_y &= \frac{y_1 - y_2}{v_{y}^1 - v_{y}^2} & b_y &= \frac{1}{2}(y_1 - v_{y}^{1}f_y + y_2 - v_{y}^{2}f_y)
\end{aligned}
\end{align*}

If we assume that the optical center is positioned at the center of the image, and the camera model adheres to the pinhole model, the estimation of the camera's intrinsic parameters can be simplified to a 1-Degree of Freedom (1-DoF) task. Specifically, we can estimate the focal length of the camera by enumerating candidate values.



\subsection{Downstream Application: 3D Reconstruction.} 
\label{sec: 3d_recon}
With the generated depth map $\hat{d}$ and estimated camera intrinsic $f$, we can reconstruct the 3D point cloud using a simple pinhole camera model.
However, the depth map we generate is the affine-invariant depth, which includes a concept of shift that leads to the distortion of the reconstructed point cloud. To mitigate this issue, we leverage the frozen shift model~\cite{yin2020learning} to recover the shift.
With the recovery of the shift in our depth map $\hat{d}$, we can obtain the point cloud $\mathbf{P}$ as:
\begin{equation}
\mathbf{P} = \hat{d} \cdot  [\frac{x - b_x}{f_x}, \frac{y - b_y}{f_y}, 1 ]^T
\label{eq:incidence_field}
\end{equation}
Here, $x$ and $y$ represent the pixel coordinates of the depth map, while the estimated $f=f_x=f_y$ and $b_x$, $b_y$ denote the location of the map's center in our pinhole camera model.

\begin{table*}
\centering
\resizebox{\linewidth}{!}{
\begin{tabular}{l | cc cc cc cc cc cc cc cc}
\hline
\multirow{2}{*}{Methods} & \multicolumn{2}{c }{NuScenes~\cite{caesar2020nuscenes}} & \multicolumn{2}{c}{KITTI~\cite{geiger2013vision}} & \multicolumn{2}{c}{CitySpace~\cite{cordts2016cityscapes}}& \multicolumn{2}{c}{NYUv2~\cite{silberman2012indoor}}& \multicolumn{2}{c}{SUN3D~\cite{xiao2013sun3d}}& \multicolumn{2}{c}{ARKitScenes~\cite{baruch2021arkitscenes}}& \multicolumn{2}{c}{Objectron~\cite{ahmadyan2021objectron}}& \multicolumn{2}{c}{MVImgNet~\cite{yu2023mvimgnet}} \\
&$e_f$ & $e_b$ & $e_f$ & $e_b$ & $e_f$ & $e_b$ & $e_f$ & $e_b$ & $e_f$ & $e_b$ & $e_f$ & $e_b$ & $e_f$ & $e_b$ & $e_f$ & $e_b$  \\
\hline
Perspective~\cite{jin2023perspective}\tiny{CVPR'23}&$0.378$&$0.286$&$0.631$&$ 0.279$&$0.624$&$ 0.316$&$0.261$&$0.348$&$0.325$&$ 0.367$&$0.260$&$  0.385$&$0.838$&$0.272$&$0.601$&$0.311$   \\
WildCamera~\cite{Wildcamere} \tiny{Nips'23}&$0.102 $&$0.087$&$0.111$&\textbf{0.078}&$0.108 $&$ 0.110$&$0.086 $&$0.174$&$0.113 $&$ 0.205$&$0.140 $&$0.243$&\textbf{0.078}&$0.070$&\textbf{0.101}&$0.081$   \\
DiffCalib (w/o depth)&\textbf{0.075}&\textbf{0.022}&\textbf{0.087}&$0.094$&\textbf{0.062}&\textbf{0.047}&\textbf{0.057}&\textbf{0.022}&\textbf{0.059}&\textbf{0.023}&\textbf{0.107}&\textbf{0.027}&$0.114 $&\textbf{0.021}&$0.108 $&\textbf{0.031}   \\
\hline
\end{tabular}
}

\caption{\small \textbf{Monocular camera calibration on the testing split of trained datasets.}
We compare our method with Perspective~\cite{jin2023perspective} and WildCamera~\cite{Wildcamere}.
Our DiffCalib trained with intrinsic only (w/o depth) outperforms the state-of-the-art methods.
}
\label{tab:train}
\end{table*}

\begin{table*}
\centering
\begin{tabular}{l | cc cc cc cc cc}
\hline
\multirow{2}{*}{Methods} & \multicolumn{2}{c }{Waymo~\cite{sun2020scalability}} & \multicolumn{2}{c}{RGBD~\cite{sturm2012benchmark}} & \multicolumn{2}{c}{ScanNet~\cite{dai2017scannet}}& \multicolumn{2}{c}{MVS~\cite{fuhrmann2014mve}}& \multicolumn{2}{c}{Scenes11~\cite{chang2015shapenet}} \\
& $e_f$ & $e_b$ & $e_f$ & $e_b$ & $e_f$ & $e_b$ & $e_f$ & $e_b$ & $e_f$ & $e_b$ \\
\hline
Perspective~\cite{jin2023perspective}\tiny{CVPR'23} & $0.444$ & $0.020$ & $0.166$ & $0.000$ & $0.189$ & $0.010$ & $0.185$ & $0.000$ & $0.211$ & $0.000$ \\
WildCamera~\cite{Wildcamere}
\tiny{Nips'23} & $0.210$ & $0.053$ & $0.097$ & $0.039$ & $0.128$ & $0.041$ & $0.170$ & $0.028$ & $0.170$ & $0.044$ \\
DiffCalib (w/o depth) & $0.188$ & $0.053$ & $0.092$ & \textbf{0.018} & $0.089$ & 0.041 & $0.135$ &  $0.032$  & \textbf{0.108} & \textbf{0.029} \\
DiffCalib & \textbf{0.145} & \textbf{0.053} & \textbf{0.084} & $0.040$ & \textbf{0.055 } & \textbf{0.036} & \textbf{0.108} & \textbf{0.036} & $0.176$  &  $0.038$  \\
\hline
WildCamera + Asm
\tiny{Nips'23} & $0.157$ & $0.020$ & $0.067$ &  $0.000$  & $0.109$ & $0.010$ & $0.127$ & $0.000$ & $0.117 $ & $0.000$ \\
DiffCalib (w/o depth) + Asm & $0.246$ & $0.020$ &  \textbf{0.052}  & $0.000$ &  $0.071$  &  $0.010$  &  $0.112$  &  $0.000$  & \textbf{0.081 } &  $0.000$  \\
DiffCalib + Asm & \textbf{0.120} &  \textbf{0.020}  & $0.062$ & \textbf{0.000} & \textbf{0.042} & \textbf{0.010} & \textbf{0.081} & \textbf{0.000} &  $0.146$  &  \textbf{0.000}  \\
\hline
\end{tabular}
\caption{\small \textbf{Monocular camera calibration on zero-shot datasets.}
We present our results with three configurations. `DiffCalib' represents the pipeline where we jointly train the incident map and the depth map. `DiffCalib (w/o depth)' represents that only the incident map is trained. `Asm' represents the assumption that the image center of the simple camera model is fixed to be the optical center, as introduced in~\cite{Wildcamere}. Compared to previous methods, our method achieves state-of-the-art performance on zero-shot datasets.}
\label{tab:test}
\end{table*}

\section{Experiments}

\subsection{Dataset and Evaluation Protocol}
\paragraph{Training Datasets}
We choose Hypersim~\cite{roberts2021hypersim} as our primary training dataset for incident map and depth map generation. 
The dataset contains 461 synthetic indoor scenes with depth information, and the ground-truth camera intrinsic parameters are $[889, 889, 512, 384]$, which is consistent across scenes.
We use 365 scenes for training, adhering to the recommended setup provided by the dataset. 
In order to increase the variety of training scenarios, we follow the baseline~\cite{Wildcamere} and adding NuScenes~\cite{caesar2020nuscenes}, KITTI~\cite{geiger2013vision}, CitySpace~\cite{cordts2016cityscapes}, NYUv2~\cite{silberman2012indoor}, SUN3D~\cite{xiao2013sun3d}, ARKitScenes~\cite{baruch2021arkitscenes}, Objectron~\cite{ahmadyan2021objectron}, and MVImgNet~\cite{yu2023mvimgnet} as our additional training data.
To ensure consistency in training across different datasets, we only consider the camera intrinsic parameters of these additional data and replace the depth input with a copied image input. 
Additionally, 
Since each image in the same dataset shares the same camera settings,
following the approach in ~\cite{lee2021ctrl}, we augment the intrinsics by randomly enlarging images up to twice of their size and then cropping them to the suitable size as our model input. This augmentation introduces variations in intrinsic parameters, addressing the scarcity of intrinsic variations within the dataset.

\paragraph{Testing Datasets}
For monocular camera calibration, our evaluation encompasses datasets such as Waymo~\cite{sun2020scalability}, RGBD~\cite{sturm2012benchmark}, ScanNet~\cite{dai2017scannet}, MVS~\cite{fuhrmann2014mve}, and Scenes11~\cite{chang2015shapenet}. We ensure alignment with the benchmark provided by WildCamera~\cite{Wildcamere} for this task.

\paragraph{Evaluation Protocol}
For camera intrinsic estimation assessment, we adhere to the prescribed evaluation protocol in~\cite{Wildcamere}, employing the metrics: 
\begin{equation}
\begin{aligned}
    e_f &= \max\left(\frac{|f_x' - f_x|}{f_x}, \frac{|f_y' - f_y|}{f_y}\right), \\
    e_b &= \max\left(2 \cdot \frac{|b_x' - b_x|}{w}, 2 \cdot \frac{|b_y' - b_y|}{h}\right)
\end{aligned}
\end{equation}
where $f_x$ and $f_y$ represent the focal lengths along the two axes, $b_x$ and $b_y$ denote the location of the optical center, $w$ and $h$ represent the width and height of the image, respectively.

\subsection{Implementation Details}
We leverage the pre-training model provided by Stable Diffusion v2.1 \cite{rombach2022high}, wherein we freeze the VAE encoder and decoder, focusing solely on training the U-Net. This training regimen adheres to the original pre-training setup with a v-objective. Moreover, we configure the noise scheduler of DDPM with 1000 steps to optimize the training process.
The training regimen comprises 30,000 iterations, with a batch size of 16. 
To accommodate the training within a single GPU, we accumulate gradients over 16 steps. 
We employ the Adam optimizer with a learning rate of $3 \times 10^{-5}$. 
Typically, achieving convergence during our training process necessitates approximately 12 hours when executed on a single Nvidia RTX A800 GPU card. We set the ensemble size as 10, meaning we aggregate predictions from 10 inference runs for each image.


\subsection{Quantitative Comparison}

\paragraph{Quantitative Comparison of Monocular Camera Calibration.}
We present the monocular camera calibration results separately for both seen and unseen datasets in Table~\ref{tab:train} and Table~\ref{tab:test}, respectively.

In Table ~\ref{tab:train}, for fair comparison, We utilize the same data as WildCamera~\cite{Wildcamere} to train our method specifically for the incident map. We then evaluate the metrics on the test split of the seen dataset. 
From Table ~\ref{tab:train}, it can be found that our approach achieves significant improvements on most of the datasets. 
For the datasets Objectron~\cite{ahmadyan2021objectron} and MVImgNet~\cite{yu2023mvimgnet}, we do not achieve the best performance in focal length estimation. 
The reason is that these datasets are focused on object-level scenes, making it challenging to predict the focal length accurately. Nevertheless, we observe significant improvements in the prediction of optical center parameters as $e_b$.

In Table ~\ref{tab:test}, our method is evaluated on zero-shot in-the-wild datasets, demonstrating its generalization ability in diverse scenes. It can been seen that our approach outperforms others, highlighting its effectiveness in real-world scenarios. Moreover, by comparing the proposed DiffCalib with DiffCalib(w/o depth), our method showcases robust joint-learning capabilities, resulting in enhancements across most datasets except RGBD~\cite{sturm2012benchmark} and Scenes11~\cite{chang2015shapenet}. 
As discussed in ~\cite{Wildcamere}, the reason may be linked to the default intrinsic parameters provided by the camera manufacturer for the RGBD~\cite{sturm2012benchmark} dataset, which lack calibration and may therefore be less accurate.
For the Scenes11~\cite{chang2015shapenet}, since it comprises images generated from randomly shaped and moving objects, which are quite perplexing even for human comprehension.
Thus, on these two datasets, only considering incident map without depth estimation achieve better results.

\paragraph{Quantitative Comparison of Zero-Shot 3D Scene Reconstruction}
To demonstrate the robustness and accuracy of our reconstruction method, we conduct a comparative evaluation against LeReS~\cite{Wei2021CVPR_leres}, a method that also utilizes a single image for 3D scene reconstruction. The comparison results are shown in Table~\ref{tab:recon_evaluation}. We employ widely used metrics in 3D reconstruction scenarios, including the Chamfer $l_1$ distance $C$-$L_1$ and $F$-$score$ with a 5cm threshold, to evaluate the point cloud.
For each method, we obtain point clouds that are aligned in scale with ground truth depth, with the pose set to the identity matrix. Subsequently, we evaluated the predicted point cloud against the ground truth point cloud, generated by projecting ground truth depth and camera intrinsic parameters. The evaluation is performed by both the $C$-$L_1$ distance and $F$-$score$ metrics.
As can be seen from Table~\ref{tab:recon_evaluation},  
our proposed approach shows great superior performance over LeReS~\cite{Wei2021CVPR_leres}, 
with a large margin of 30\% improvement in the $F$-$score$ metric. 

By visualizing the intrinsic characteristics of ScanNet~\cite{dai2017scannet}, we observe discrepancies between the predicted and ground truth focal lengths. Specifically, while LeReS predicts a focal length of approximately 980, our method predicts a value closer to 1100, with the GT focal length is 1165.72. This discrepancy results in improved results with our method, as evidenced by the evaluation metrics. As for $C-L1$, we find the generation depth failed in some specific condition and cause the worse results. This failure may be attributed to the limited utilization of depth information.

\begin{table}
    \centering
        \begin{tabular}{l | cc cc}
        \hline
        \multirow{2}{*}{Methods} & \multicolumn{2}{c}{NYU[1]} & \multicolumn{2}{c}{ScanNet~\cite{dai2017scannet}} \\
        & $C$-$l_1 \downarrow$ & $F$-$score \uparrow$ & $C$-$l_1 \downarrow$ & $F$-$score \uparrow$ \\
        \hline
    LeReS~\cite{yin2020learning} & $0.145$ & $0.333$ & $\textbf{0.126}$ & $0.377$ \\
        DiffCalib & $\textbf{0.127}$ & $\textbf{0.433}$ & $0.150$ & $\textbf{0.508}$ \\
        \hline
        \end{tabular}
    \caption{Qualitative comparison of 3D reconstruction from a single image. We compare the reconstruction performance with LeReS \cite{yin2020learning} on two zero-shot datasets: NYU~\cite{silberman2012indoor} and ScanNet~\cite{dai2017scannet}. The chamfer distance $C$-$L_1$ and $F$-$score$ metrics with a threshold of 5cm are evaluated here.}
    \label{tab:recon_evaluation}
\end{table}

\subsection{Ablation Study}
\paragraph{Pre-train of U-Net.} 
Table \ref{tab:pre-train} illustrates the performance of camera intrinsic estimation with and without pre-training of U-Net. To ensure the network's fitting, we train it for 30k iterations, which is 10k more than using pre-training, and test it on the validation data to ensure its fitting. It is evident that the performance can be significantly improved with the pre-training of U-Net. This underscores the importance of employing a generalizable and strong prior for other works in this domain. We only list the worst results of the two datasets, which can largely represent the situation of U-Net pre-training.

\begin{table}
    \centering
    \begin{tabular}{l |cc cc}
    \hline
    \multirow{2}{*}{Methods} & \multicolumn{2}{c}{Waymo~\cite{sun2020scalability}} & \multicolumn{2}{c}{Scenes11~\cite{chang2015shapenet}} \\
    &$e_f$ & $e_b$ & $e_f$ & $e_b$ \\
    \hline
    DiffCalib (w/o pre-trained) & $0.263$ & $0.062$ & $0.214$ & $0.034$ \\
    DiffCalib   & \textbf{0.188} & \textbf{0.054} & \textbf{0.108} & \textbf{0.029} \\
    \hline
    \end{tabular}
    \caption{\textbf{Ablation study for the pre-trained parameters of the U-Net model. We observe that the U-Net pre-training parameters can enhance the performance and facilitate the convergence.}}
    \label{tab:pre-train}
\end{table}

\paragraph{Ensemble of Incidence Noise.} 
Using the diffusion model to generate the incident map offers a significant advantage in terms of variety. By denoising the multiple noise and averaging the corresponding results, the confidence of the predicted incident map increases. The results in Table~\ref{tab:ensemble_noise} demonstrate this benefit, showing that ensemble noise of the incident map leads to better results in several in-the-wild datasets. This aligns with real-world scenarios where images with slight variations may share the same incident map.
\begin{table}
    \resizebox{0.48\textwidth}{!}{
        \begin{tabular}{l | c | cc cc cc}
        \hline
        \multirow{2}{*}{Methods} & Ensemble &  \multicolumn{2}{c}{Waymo~\cite{sun2020scalability}} & \multicolumn{2}{c}{RGBD~\cite{chang2015shapenet}} & \multicolumn{2}{c}{MVS~\cite{fuhrmann2014mve}} \\
        & Times &$e_f$ & $e_b$ & $e_f$ & $e_b$ & $e_f$ & $e_b$ \\
        \hline
        w/o ensemble & 0 &  $0.160$ & $0.098$ & $0.141$ & $0.068$ & $0.224$ & $0.077$ \\
        DiffCalib  & 10 & \textbf{0.145} & \textbf{0.053} & \textbf{0.084} & \textbf{0.040} & \textbf{0.108} & \textbf{0.036} \\
        \hline
        \end{tabular}
    }
    \captionof{table}
    {\textbf{Ablation study of the ensemble process.} 
    With the ensemble process, our DiffCalib achieves much better robustness on camera intrinsic estimation. This proves the necessity and advantage of reformulating monocular camera calibration as a generation task. The generation paradigm together with the ensemble process improves the estimation confidence significantly.
    }
    \label{tab:ensemble_noise}
\end{table}

\paragraph{The Utilization of Incident Maps Improves Depth Estimation.} 
For depth estimation, we conduct testing on the NYU~\cite{silberman2012indoor} and ETH3D ~\cite{schops2017multiEth3d} datasets, following the settings of affine-invariant depth evaluation. Given that our training data comprises solely indoor scenes, we evaluate the model on indoor scenes from the NYU~\cite{silberman2012indoor} dataset and both indoor and outdoor scenes from the ETH3D~\cite{schops2017multiEth3d} dataset. 
As shown in Table~\ref{tab:incidence_benefit_depth}, incorporating both depth and incident maps not only enhances the accuracy of intrinsic estimation but also improves depth estimation performance. A comparison between our generation model trained with depth alone and the model generating both depth and incident map reveals that the joint incorporation of the incident map leads to superior results in depth estimation. Both models are trained on the same dataset, Hypersim, with the sole distinction being the supervision of depth and joint incident map. We ensured the alignment of our evaluation with other depth estimation methods to validate the effectiveness of the results.

\begin{table}
    \centering
    \begin{tabular}{l| cc cc}
    \hline
    \multirow{2}{*}{Methods} & \multicolumn{2}{c}{NYU~\cite{silberman2012indoor}(indoor)} & \multicolumn{2}{c}{ETH3D~\cite{schops2017multiEth3d}} \\
    &$AbsRel \downarrow$ & $\delta1 \uparrow$ & $AbsRel \downarrow$ & $\delta1 \uparrow$ \\
    \hline
    DiffCalib (w/o incident) & $0.086$ & $0.924$ & $0.101$ & $0.903$ \\
    DiffCalib   & \textbf{0.082} & \textbf{0.927} & \textbf{0.086} & \textbf{0.918} \\
    \hline
    \end{tabular}
    \caption{\textbf{Ablation for the influence of incident map estimation to depth estimation. 
    } `DiffCalib (w/o incident)' refers to training only with depth maps, whereas `DiffCalib' means training with both depth and incident maps. It is worth mentioning that we solely utilize the Hypersim dataset, hence the baseline results may not be outstanding compared to the state-of-the-art.}\label{tab:incidence_benefit_depth}
\end{table}

\begin{figure*}
	\centering 
 \includegraphics[width=1\textwidth]{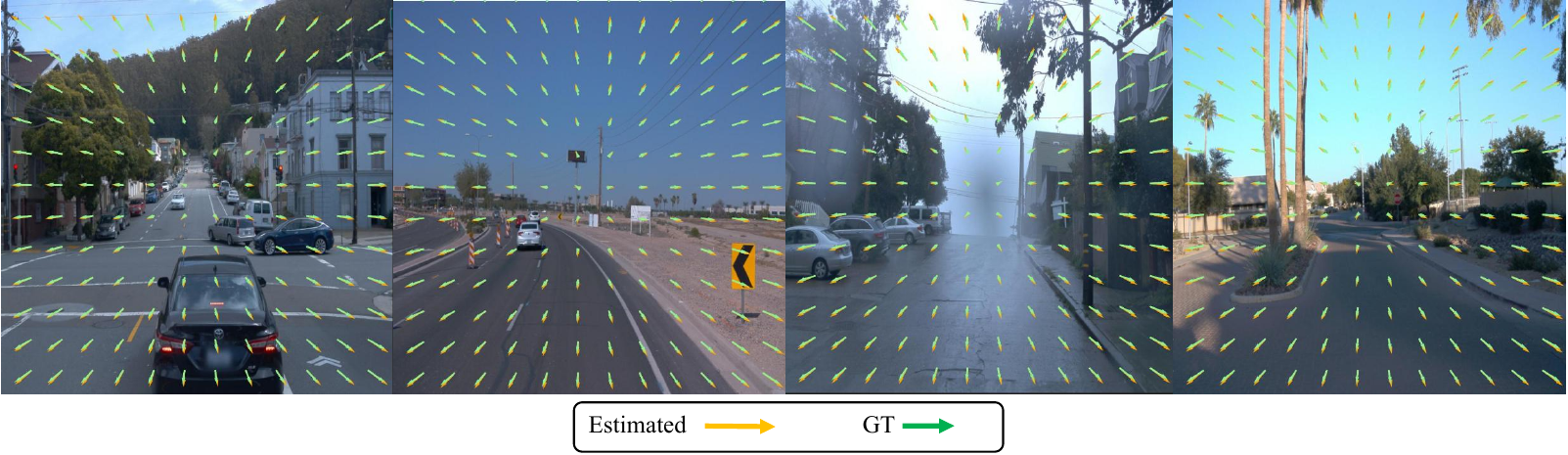}
	\caption{Visualization of the estimated and ground-truth (GT) incident map on the outdoor dataset Waymo~\cite{sun2020scalability}.}
	\label{fig:incidence_waymo}
\end{figure*}
\begin{figure*}[!t]
	\centering 
	\includegraphics[width=1\textwidth]{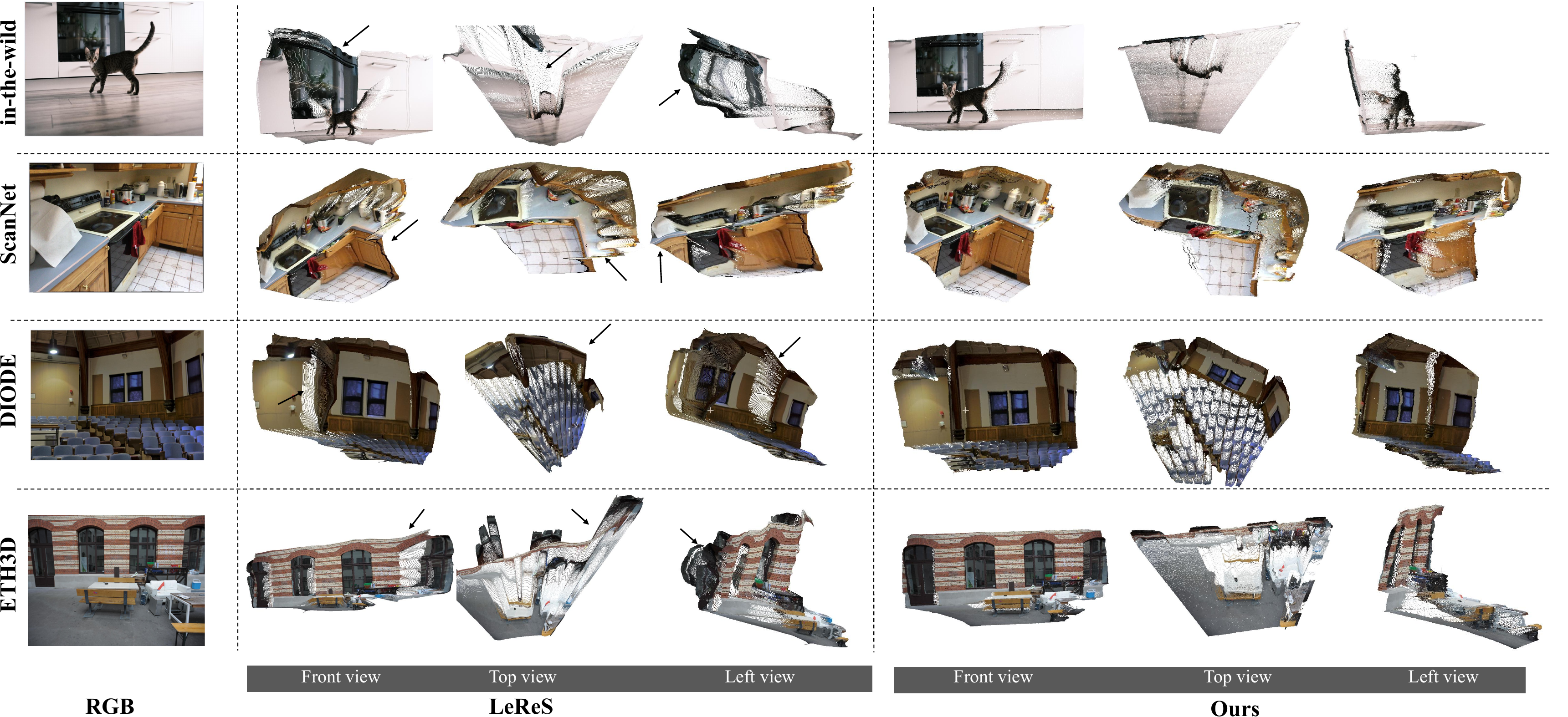}
	\caption{Qualitative comparison of 3D reconstruction. We compare with LeReS~\cite{yin2020learning} across diverse scenes.}
	\label{fig:recon_dataset}
\end{figure*}

\subsection{Visualization Comparison}
\paragraph{Qualitative Analysis of the Incident Map.}
We compare our generation incident map with the ground-truth (GT) incident map. The comparison results are shown in Figure~\ref{fig:incidence_waymo}. Remarkably, the generated incident map exhibits a distinct directionality and has little difference from the ground-truth (GT) incident map. It is further intuitively demonstrated that our approach can achieve robust and accurate calibration estimation through the generation of an incident map.




\paragraph{Qualitative Comparison with 3D Reconstruction.}
We conduct a comparative analysis in Figure~\ref{fig:recon_dataset} between our single image 3D reconstruction method and the recent LeReS approach \cite{yin2020learning} on ScanNet, DIODE, ETH3D dataset, and in-the-wild condition, both of which rely solely on a single image for 3D reconstruction. 
our method consistently outperforms LeReS in terms of both detail preservation and geometric accuracy.



\section{Conclusion}
In this paper, we have presented a diffusion-based approach, namely DiffCalib, for monocular camera calibration with a single in-the-wild image.
We reformulate the calibration problem as a diffusion-based dense incident map generation task. By leveraging the successful visual prior knowledge transferring from image generation to incidence estimation and depth estimation, our approach enables more robust and accurate in-the-wild camera calibration, showing superior generalization ability and effectiveness upon existing methods. 
Beyond calibration, we show that the robust performance of our DiffCalib can greatly benefit downstream applications such as 3D scene reconstruction from a single image.


\bibliographystyle{ACM-Reference-Format}
\bibliography{sample-base}

\newpage
\twocolumn
\appendix
\newpage

\section*{\Large\textbf{Supplementary Material}}

\section{EXPERIMENT}
\subsection{Experiment Details}
\paragraph{Datasets for Training}
To train our diffusion-based DiffCalib model which utilizes a single in-the-wild image for monocular camera calibration, we collect a variety of datasets. Note that we only train our model with depth on Hypersim~\cite{roberts2021hypersim} dataset.
\begin{itemize}
    \item Hypersim~\cite{roberts2021hypersim} is a synthetic but photorealistic dataset with 461 indoor scenes. We use the official split with around 54K samples from 365 scenes for training and the camera intrinsic is [889, 889, 512, 384].
    \item NuScenes~\cite{caesar2020nuscenes}, KITTI~\cite{geiger2013vision}, CitySpace~\cite{cordts2016cityscapes} are all autonomous driving datasets, and in all of the scenes, we use the camera's intrinsic accurate calibration with a checkerboard. The camera intrinsic of each dataset is [1266.24, 1266.42, 816.27, 491.51], [718.86, 718.86, 607.19, 185.22] and [2267.86, 2230.28, 1045.53, 518.88].
    \item NYUv2~\cite{silberman2012indoor}, SUN3D~\cite{xiao2013sun3d}, ARKitScenes~\cite{baruch2021arkitscenes} are all indoor datasets and in all of the scenes, we use the camera's intrinsic accurate calibration with a checkerboard. The camera intrinsic of each dataset is [518.85, 519.47, 325.58, 253.74], [570.34, 570.32, 320.00, 240.00] and [1601.95, 1601.95, 936.55, 709.61].
    \item Objectron~\cite{ahmadyan2021objectron} is an object-centric dataset and in all of the scenes, we use the camera's intrinsic accurate calibration with a checkerboard and the camera intrinsic is [1579.18, 1579.18, 721.01, 934.70].
    \item MVImgNet~\cite{yu2023mvimgnet} is a real-world objects dataset. The camera intrinsic is varying and we use the camera's intrinsic which is computed via an SfM method.
\end{itemize}

\paragraph{Datasets for Testing}
To evaluate our model's generalization in the unseen scenes, we utilize a variety of datasets that are not included in training data.
\begin{itemize}
    \item Waymo~\cite{sun2020scalability} is an autonomous driving dataset and in all of the scenes, we use the camera's intrinsic accurate calibration with a checkerboard and the camera intrinsic is [2060.56, 2060.56, 947.46, 634.37].
    \item RGBD~\cite{sturm2012benchmark} is an indoor dataset that the intrinsic is provided by the camera manufacturer without a calibration process. The camera intrinsic is [570.00, 570.00, 320.00, 240.00].
    \item ScanNet~\cite{dai2017scannet} is an indoor dataset and in all of the scenes, we use the camera's intrinsic accurate calibration with a checkerboard and the camera intrinsic is [1165.72, 1165.74, 649.09, 484.77].
    \item MVS~\cite{fuhrmann2014mve} is a hybrid dataset and the intrinsic is provided by the camera manufacturer without a calibration process. The camera intrinsic is [570.00, 570.00, 320.00, 240.00].
    \item Scenes11~\cite{chang2015shapenet} is a synthetic dataset and the images are generated from randomly shaped and moving objects. The camera manufacturer provides the camera intrinsic without a calibration process as [570.00, 570.00, 320.00, 240.00].
\end{itemize}

\paragraph{Training Details}
We train our model using three different configurations adapted to specific tasks.
Firstly, for our model without depth, which serves as the baseline comparison to the benchmark~\cite{Wildcamere}, we utilize training datasets excluding Hypersim.
Next, we utilize all available training datasets to train our model with depth, while not supervising the depth map when it is not part of the input.
Finally, for depth estimation and 3D reconstruction tasks, we solely rely on the Hypersim dataset for training, incorporating both depth map and incident map


\subsection{Additional Experiments}
\paragraph{The Validity of Geometry.}
Since depth information can enhance the prediction of the incident map, we investigate whether normal information, which also belongs to geometry information, can offer similar benefits. Initially, we note that the incident map's semantic level of detail is limited, and its per-pixel values are relative due to its derivation from the 4-degree-of-freedom camera intrinsic parameters. Consequently, we hypothesize that leveraging dense and geometric normal maps may also improve the prediction of the incident map.
In Table~\ref{tab:normal-benefit-incident}, we observe significant improvements across several in-the-wild datasets. We employ a method similar to the depth joint incident map but substituting depth with normal information. This approach yields promising results, suggesting that geometric information can indeed enhance the prediction of the incident map. Note that we use the same setting as Table 2 on page 6 of the main paper.
\begin{table*}[t!]
\vspace{2mm}
\centering
\begin{tabular}{l cc cc cc cc cc}
\hline
\multirow{2}{*}{Methods} & \multicolumn{2}{c }{Waymo~\cite{sun2020scalability}} & \multicolumn{2}{c}{RGBD~\cite{sturm2012benchmark}} & \multicolumn{2}{c}{ScanNet~\cite{dai2017scannet}}& \multicolumn{2}{c}{MVS~\cite{fuhrmann2014mve}}& \multicolumn{2}{c}{Scenes11~\cite{chang2015shapenet}} \\
& $e_f$ & $e_b$ & $e_f$ & $e_b$ & $e_f$ & $e_b$ & $e_f$ & $e_b$ & $e_f$ & $e_b$ \\
\hline
DiffCalib (w/o normal) & $0.246$ & \textbf{0.020} & 0.052  & \textbf{0.000} &  $0.071$  &  $0.010$  &  \textbf{0.112}  &  \textbf{0.000}  & \textbf{0.081 } &  \textbf{0.000}  \\
DiffCalib (w normal) & \textbf{0.171} &  \textbf{0.020}  & \textbf{0.047} & \textbf{0.000} & \textbf{0.041} & \textbf{0.010} & 0.132 & \textbf{0.000} &  $0.091$  &  \textbf{0.000}  \\
\hline
\end{tabular}
\caption{\small \textbf{Ablation of geometry information surface normal contributes to monocular camera calibration.}
We present our results with two configurations. ‘DiffCalib(w normal)' represents the pipeline where we jointly train the incident and surface normal maps. 'DiffCalib (w/o normal)' represents that only the incident map is trained. Both results are based on the assumption that the image center of the simple camera model is fixed to be the optical center, as introduced in~\cite{Wildcamere}.
}
\label{tab:normal-benefit-incident}
\end{table*}

\begin{figure*}
	\centering 
	\includegraphics[width=0.85\textwidth]{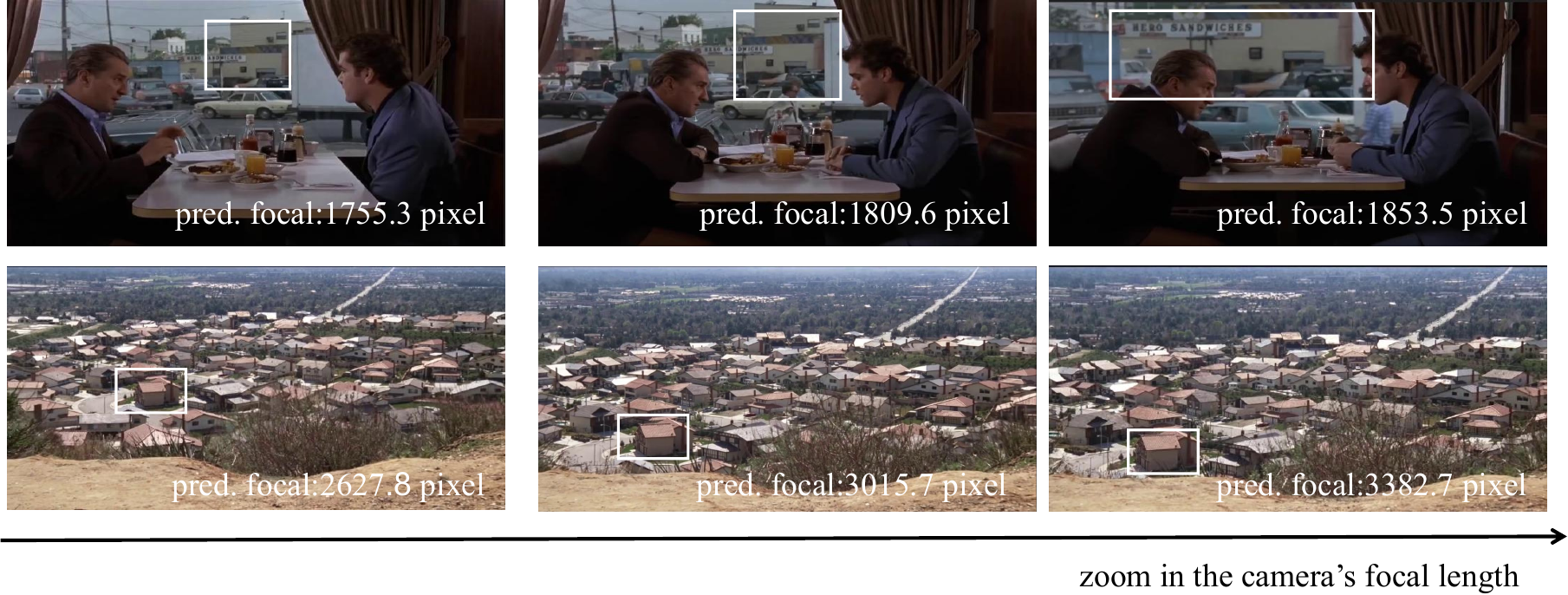}
	\caption{Visualization of zooming in camera focal length on in-the-wild scenes. We present images with increasing camera focal lengths from left to right, with our predicted focal length displayed in the bottom right corner. We notice a notable increase in the estimated focal length as we zoom in on the camera's focal length.}
	\label{fig:vis_focal}
\end{figure*}

\paragraph{Visualization of Focal Length.}
Due to the exceptional generalization performance exhibited by our model across unseen datasets, we employ our model to predict the focal length of images sourced from the Internet, as illustrated in Figure~\ref{fig:vis_focal}. We observed the trend of focal length change. In the first row, we depict indoor scenes, whereas the camera focal length increases, the estimated focal length also rises, indicating our model's ability to perceive changes in focal length against varying backgrounds.
Moving to the second row, we illustrate distant views. By zooming in the camera focal, we observe a significant increase in the estimated focal length, providing further evidence of the robust generalization capabilities of our model.

Additionally, in the supplementary video file "estimation of focal length.mp4," we demonstrate how the model predicts the focal length and its changes corresponding to variations in camera focal length. This further illustrates the robustness of our model. Please note that we utilize polyfit to fit this curve and visualize the changing trend of focal length.

\subsection{Illustration of Incident Map and Depth Map}
\begin{figure*}
	\centering 
	\includegraphics[width=1\textwidth]{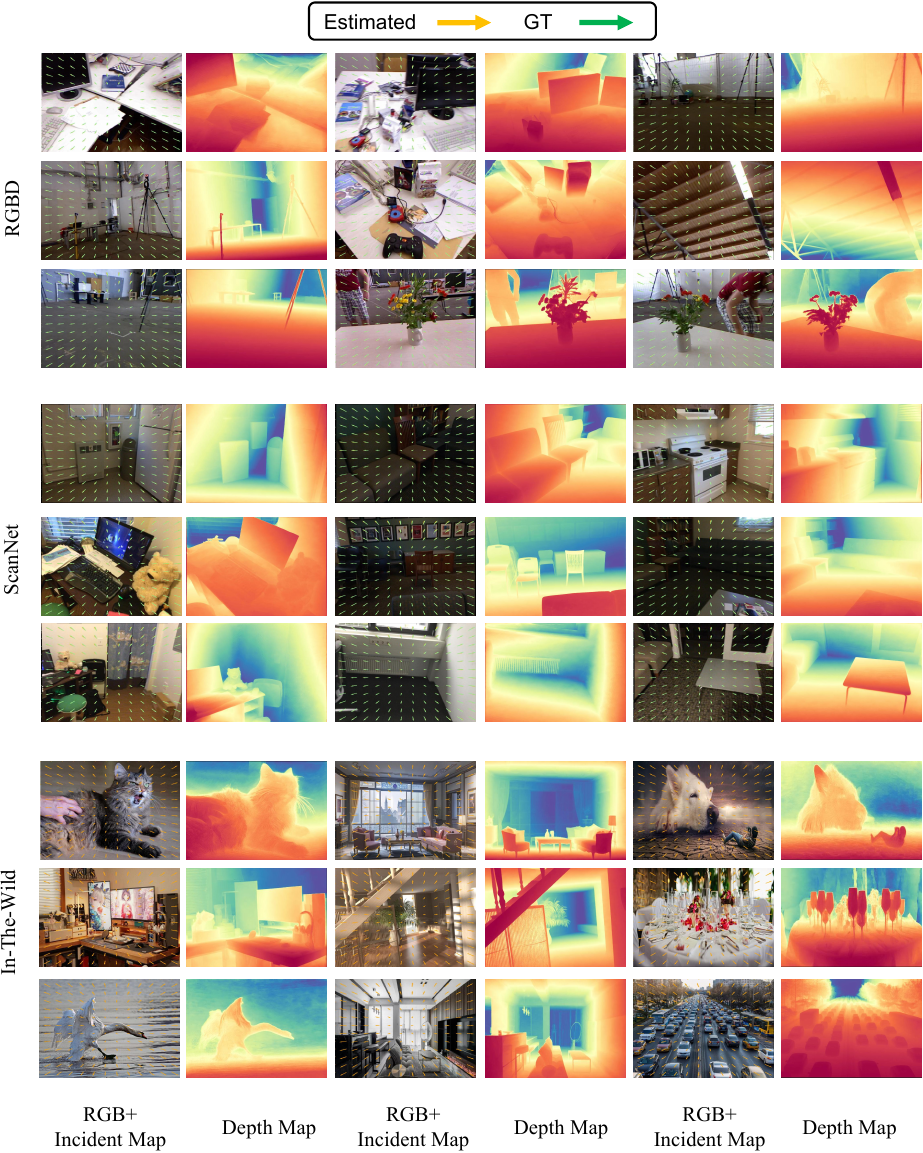}
	\caption{Visualization of incident map and depth map on zero-shot datasets and in-the-wild scenes. Note that with in-the-wild conditions we do not have GT incident maps.}
	\label{fig:supp_map}
\end{figure*}

We illustrate some examples of generation incident map and depth map in the zero-shot dataset RGBD~\cite{sturm2012benchmark} and ScanNet~\cite{dai2017scannet} in Figure~\ref{fig:supp_map}. Besides, we also display some examples which are in-the-wild scenes.

\subsection{Illustration of the Reconstructed Point Cloud}
\begin{figure*}
	\centering 
	\includegraphics[width=1\textwidth]{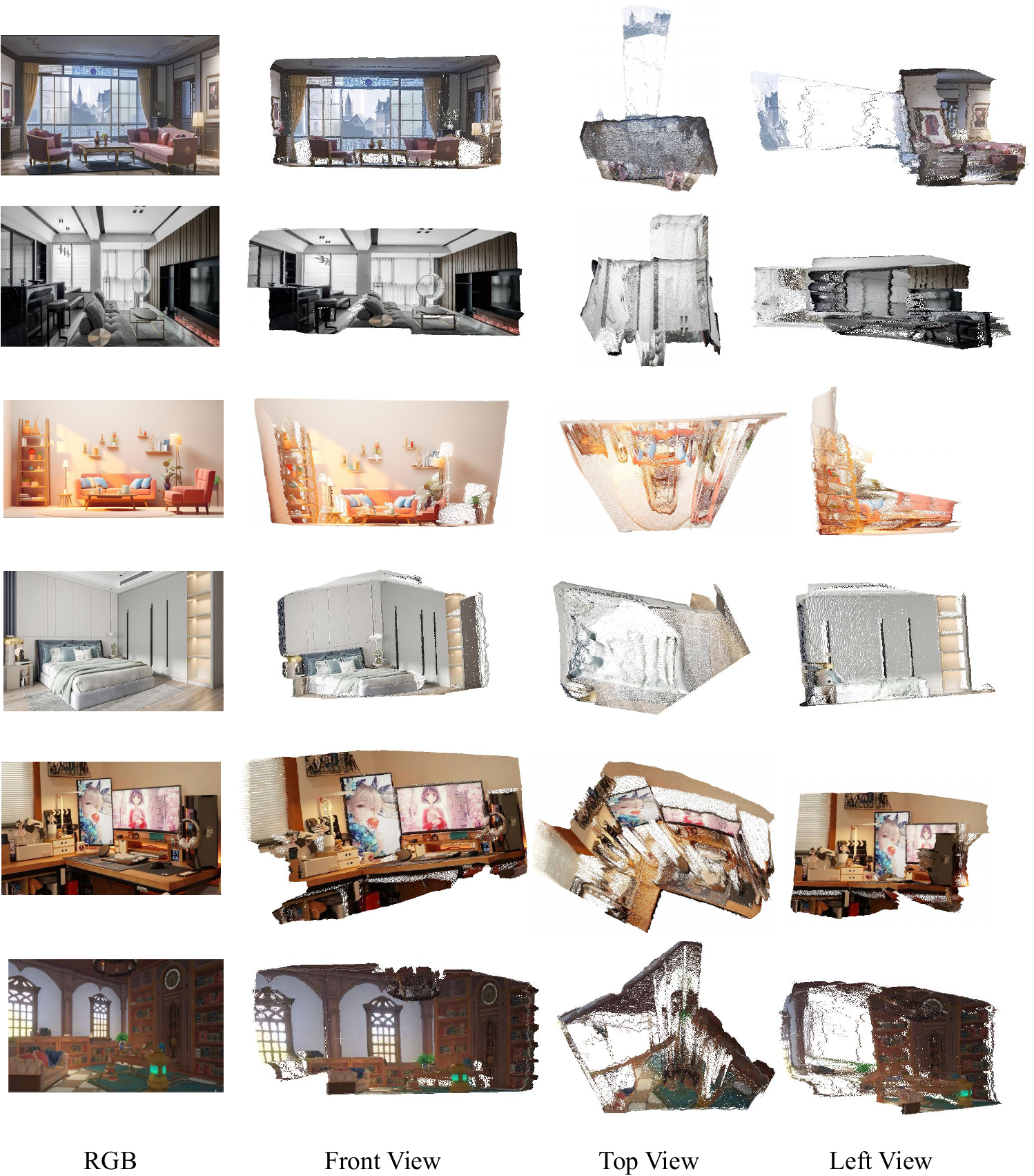}
	\caption{Visualization of point cloud on in-the-wild scenes. The first column is the RGB images and the next three columns are the front view, top view, and left view, respectively.}
	\label{fig:supp_point_cloud}
\end{figure*}

We illustrate some examples of the reconstructed 3D point cloud from our DiffCalib model in Figure~\ref{fig:supp_point_cloud}. All the examples are from the Internet and there are even some virtual scenes. This shows that our method demonstrates good generalizability in in-the-wild scenes.


\end{document}